\documentclass[letterpaper, 10 pt, conference]{ieeeconf}  % Comment this line out if you need a4paper

\IEEEoverridecommandlockouts                              % This command is only needed if 
\usepackage{amsmath} % assumes amsmath package installed
\usepackage{amssymb}  % assumes amsmath package installed
\usepackage{caption}
\usepackage{cite}
\usepackage{graphics} % for pdf, bitmapped graphics files
\usepackage{graphicx}
\usepackage{subcaption}
\usepackage{multirow}

\title{\LARGE \bf
Lighthouses and Global Graph Stabilization: Active SLAM for Low-compute, Narrow-FoV Robots
}

\author{$^{1}$Mohit Deshpande, Richard Kim, Dhruva Kumar, Jong Jin Park, Jim Zamiska% <-this % stops a space
\thanks{$^{1}$All authors are with Amazon Lab126.}\\
\thanks{\tt\scriptsize{\{deshmohi,richk,dhruvkm,jongpark,jzamiska\}@amazon.com}}\\
\thanks{All authors contributed equally and are listed alphabetically.}
}

\begin{document}

\maketitle
\thispagestyle{empty}
\pagestyle{empty}

\begin{abstract}
Autonomous exploration to build a map of an unknown environment is a fundamental robotics problem. However, the quality of the map directly influences the quality of subsequent robot operation. Instability in a simultaneous localization and mapping (SLAM) system can lead to poor-quality maps and subsequent navigation failures during or after exploration. This becomes particularly noticeable in consumer robotics, where compute budget and limited field-of-view are very common.
In this work, we propose (i) the concept of \textit{lighthouses}: panoramic views with high visual information content that can be used to maintain the stability of the map locally in their neighborhoods and (ii) the final stabilization strategy for global pose graph stabilization. We call our novel exploration strategy SLAM-aware exploration (SAE) and evaluate its performance on real-world home environments.
\end{abstract}

\section{Introduction}
For many types of tasks performed by an autonomous mobile robot, especially in an indoor environment, it is often useful to have a map of the environment where the robot operates. However, in many environments, especially homes, it is often not feasible or practical to construct this map by hand since it can be time-consuming or prone to human error.

The task of autonomous exploration is to construct this map without supervision. Traditionally, most exploration strategies\cite{bircher2016receding,dang2019graph,dharmadhikari2020motion,yamauchi1997frontier} use a discretized 2D grid map or 3D voxel map where each cell/voxel has a value that represents a probability of being occupied, and they produce a plan to navigate the device to an area where it can convert unknown cells/voxels into free or occupied cells/voxels. These strategies\cite{bircher2016receding,dang2019graph,dharmadhikari2020motion} often information-theoretic to explore the map quickly but do not consider the quality of localization during exploration.

However, the quality of the final map depends on the quality of localization during exploration. The pose estimates from a Simultaneous Localization and Mapping (SLAM) system drift while navigating in an environment, and this is especially true for visual SLAM (vSLAM) systems when they encounter environments with a lack of visual features. Enough drift causes the quality of the map to degrade to the point where exploration and navigation fail. This drift requires correction in the form of loop closures. Active SLAM aims to keep the SLAM uncertainty bounded by intentionally creating loop closures using an active loop closure (ALC) module which navigates the robot back to known areas to get a loop closure to reduce uncertainty. In a pose graph SLAM formulation\cite{grisetti2010tutorial}, this corresponds to adding a constraint between two non-adjacent poses with the error between the two. However, these techniques, along with other passive SLAM strategies, often use wide FoV sensors\cite{hu2020voronoi,ramezani2020online,stachniss2004exploration}, such as 360${}^\circ$ LiDAR or range finders, or use large compute modules\cite{hu2020voronoi,lehner2017exploration}.

\begin{figure}
    \centering
    \includegraphics[width=\columnwidth]{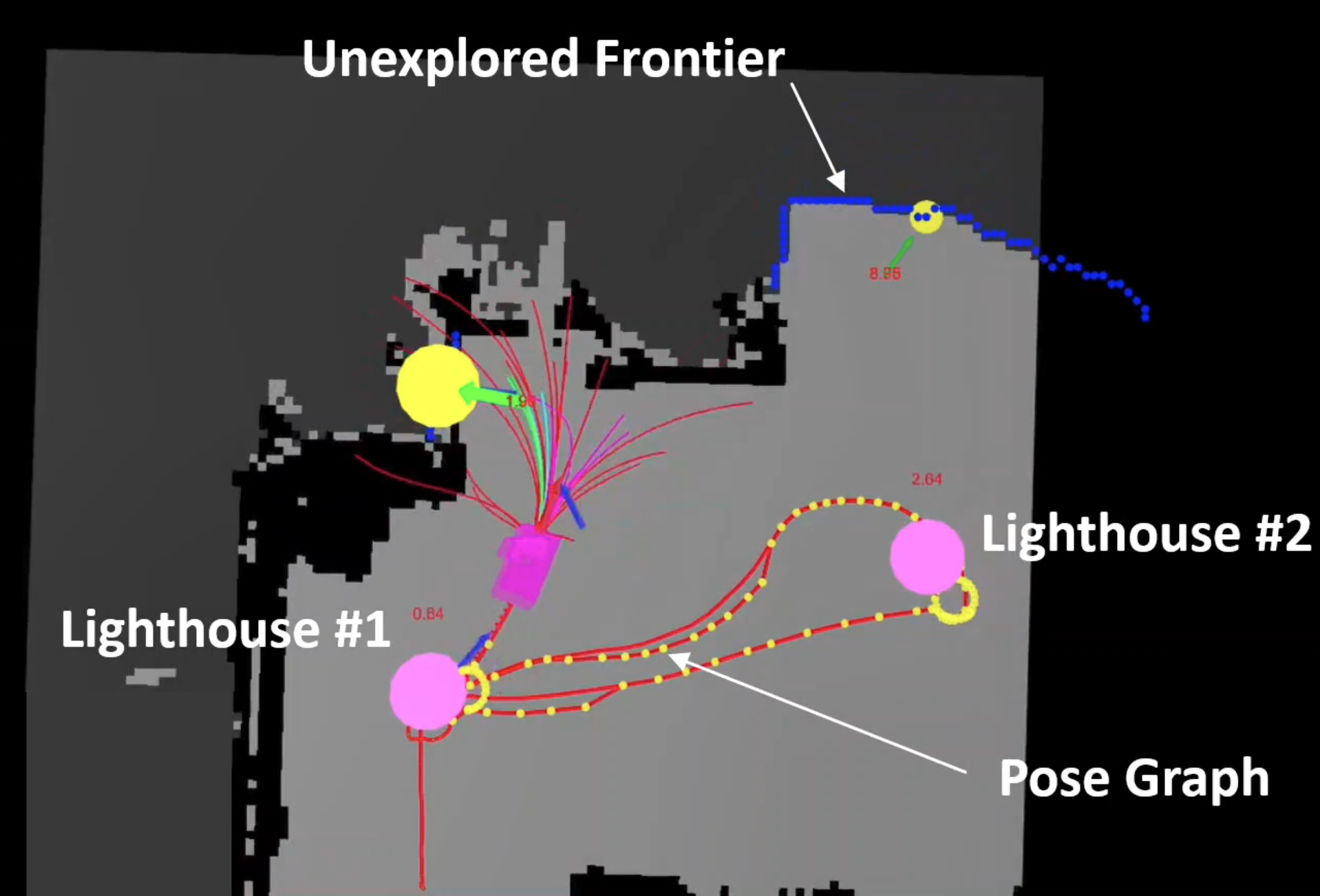}
    \caption{\small{\textbf{SLAM-aware Exploration (SAE)}. Our exploration strategy alternates between exploring frontiers, creating lighthouses, and revisiting a lighthouse. Blue points indicate frontiers points (the boundary between unoccupied cells and cells with unknown occupancy); green arrows and yellow circles indicate frontier viewpoints and their associated frontiers (larger means active). The purple box is the robot, with emanating colored lines indicating candidate local planner trajectories. Pink spheres are lighthouses which are made of spatially-clustered keyframes (small yellow rings) in the pose graph. After we create the ``home'' lighthouse near the origin, we connect each subsequently created lighthouse to a previously-created one.}}
    \label{fig:lh-alc}
\end{figure}

In this work, we introduce a new type of active SLAM exploration called \textbf{SLAM-aware Exploration (SAE)} for narrow-FoV visual SLAM and low-compute devices that creates a map while maintaining SLAM stability. Figure \ref{fig:lh-alc} shows lighthouses created by SAE and the corresponding pose graph. We evaluate our approach on a mobile robot in a home environment and show we achieve a higher exploration success rate as well as more loop closures on post-exploration navigation tasks compared to other exploration algorithms. Our major contributions are
\begin{enumerate}
    \item A structure called a \textbf{lighthouse}, which is a visually-informative location with a panoramic view, that we create and travel back to during exploration as a part of our lighthouse-based active loop closure (LH-ALC) planner to stabilize the pose graph locally during exploration.
    \item A final \textbf{Global Graph Stability (GGS)} planner that performs post-exploration pose graph stabilization.
    \item An overall system design that alternates between exploring the map and traveling back to lighthouses to bound SLAM uncertainty.
\end{enumerate}

%All of these novel contributions of our active SLAM approach are currently deployed on a mass-market, consumer-grade home robot: Amazon Astro. The experiments are also all validated on the exact same consumer robot without any additional compute modules or sensors. Active SLAM is an area in an early development stage with very few reports of field experiments\footnote{We found 52,000 patents and research work for the SLAM compared to only 150 patents and research for Active SLAM. The queries ``simultaneous localization and mapping'' and ``active simultaneous localization and mapping'' were run on Google Scholar and included patents.}\cite{placed2022survey} and no consumer-grade deployment because of the high compute and sensor cost; to the best of our knowledge, Amazon Astro is the first consumer-grade robot with a deployed functioning active SLAM.

\section{Related Work}
There have been a number of works in the active SLAM and active perception communities using a wide variety of sensors, SLAM algorithms, uncertainty metrics, and exploration termination criteria. This section groups these works by theme and compares them to our work. \cite{placed2022survey} provides a more detailed and comprehensive review.

\subsection{Rule-based Strategies}
Rule-based active SLAM strategies decompose the problem into (i) candidate action generation, (ii) utility computation, and (iii) action selection. For (i), random-sampling goal-selection strategies \cite{gonzalez2002navigation,tovar2006planning} are very compute-efficient per candidate action but not exploration efficient. Frontier-based strategies \cite{yamauchi1997frontier,keidar2012robot,quin2021approaches,holz2010evaluating,keidar2014efficient,wu2019autonomous} use the insight that areas to explore in the map are at the boundary between known and unknown space; these approaches drive the robot directly to unexplored areas of the map but are more expensive since they require searching through the map. For (ii), traditional utility functions consider distance to the candidate goals\cite{yamauchi1997frontier,holz2010evaluating,tovar2006planning}. More complex utility functions use information theory to approximate the expected reduction in uncertainty of the map at that candidate goal and balance that against the travel distance\cite{bonetto2022irotate,wu2019autonomous,gonzalez2002navigation}. To estimate the uncertainty of the SLAM part of the map, some use various optimality criteria from the Theory of Experimental Design (TOED)\cite{placed2020deep,suresh2020active}. Some work\cite{placed2022explorb,kim2015active} creates ``virtual'' edges in a pose graph and then estimates the expected uncertainty along the path to the goal. For (iii), action selection performs some kind of optimization, the simplest being greedy selection, that minimizes either covariance or entropy\cite{yamauchi1997frontier,holz2010evaluating,tovar2006planning,wu2019autonomous,gonzalez2002navigation}.

Our approach fits into this set of strategies. Our candidate action generation is frontier-based and uses the frontiers to generate candidate viewpoints. We propose a novel compute-efficient utility computation accounting for the robot dynamics, and our greedy action selection works well in practice.

\subsection{Belief-space Strategies}
Instead of operating in a discrete space, belief-space strategies optimize the continuous trajectory and require a continuous utility function. These approaches incorporate an approximate uncertainty reduction term in the optimization to prefer trajectories that move towards the goal while reducing map uncertainty\cite{chen2020active,leung2006active}.

\subsection{Deep Reinforcement Learning Strategies}
Many deep reinforcement learning approaches \cite{tai2016mobile,zhelo2018curiosity,hu2020voronoi,oh2019learning,placed2020deep} map sensor inputs, e.g., depth images and laser scans, perhaps along with some auxiliary information, into a fixed set of actions, e.g., go forward $0.2m$, turn left 8${}^\circ$, and turn right 8${}^\circ$; for these works, the reward functions consist of an extrinsic reward that performs collision avoidance and an intrinsic reward that encourages exploring new areas while minimizing uncertainty using information theory or TOED.

%These approaches often require millions of examples to produce a good exploration policy, and the runtime inference can expensive, depending on the model size. Our approach does not use any learning-based system but operates on a higher-level of abstraction using frontiers and lighthouses rather than raw sensor inputs.

\section{SLAM-aware Exploration (SAE)}

We factor the active SLAM exploration problem into two coupled parts: (i) constructing a high-fidelity 2D occupancy map that contains information about obstacles and free space for navigation and (ii) maintaining a stable SLAM pose graph. For navigation and planning, we use a 2D occupancy grid map $\mathcal{M}$ where each cell $m_{i,j}$ at index $i,j$ represents the log odds probability that cell is occupied. Note that $m_{i,j}=0$ is a special value meaning ``unknown''. Practically, this is realized using a standard depth sensor measurement model and Bayesian updates. The objective of this part of SAE is to ensure completeness of the occupancy map, i.e., there are no unknown cells in the occupancy map.

%For our SLAM formulation, we use a keyframe-based visual-inertial-odometry (VIO) SLAM pose graph formulation\cite{dellaert2017factor} shown in Equation~\ref{eq:slam}.
For our SLAM formulation, we use a keyframe-based pose graph vSLAM system\cite{dellaert2017factor}.
%\begin{align}\label{eq:slam}
%    \mathcal{X}^\star &= \underset{\mathcal{X}}{\arg\!\min}
%    \Bigg(\underbrace{\displaystyle\sum_i||f(\hat{x}_{i-1},\hat{x}_i)-x_i||^2_{\Psi_i}}_{\text{odometry factors}}\nonumber\\
%    &+\underbrace{\displaystyle\sum_{i,j}||g(\hat{x}_i,\hat{x}_j)-\ell_{i,j}||^2_{\Phi_{i,j}}}_{\text{loop closure factors}}\nonumber\\
%    &+\underbrace{\displaystyle\sum_{i}||h(\hat{x}_i)||^2_{\Omega_i}}_{\text{priors}}\Bigg)
%\end{align}
%We use $f(\cdot)$, $g(\cdot)$, and $h(\cdot)$ to represent our measurement functions for odometry, loop closures, and priors, respectively. The respective covariances are $\Psi_i$, $\Phi_{i,j}$, and $\Omega_i$. $||x||^2_\Sigma$ denotes the squared Mahalanobis distance $x^T \Sigma^{-1}x$. 
The objective of the graph stabilization part of SAE is to keep the global uncertainty of SLAM, i.e., uncertainty of the entire pose graph, bounded by (i) creating lighthouses and traveling back to them during exploration and (ii) performing a final GGS plan to create more keyframes and loop closure constraints that lower the global uncertainty. SAE maintains both a stable occupancy map and a stable pose graph as the robot \textit{incrementally} explores the environment.

SAE consists of a number of different planners working together to efficiently and stably explore the environment. A frontier-based exploration planner analyzes the occupancy map for areas of information gain and produces plans to travel to those areas. The LH-ALC planner proactively creates lighthouses at visually-informative places and monitors pose uncertainty to determine if we need to travel back to a lighthouse to locally stabilize the SLAM pose graph. Finally, after the occupancy map is fully known, the GGS planner produces plans to globally stabilize the graph.

\subsection{Frontier-based Exploration (FE)}
To construct the occupancy map, we use a frontier-based exploration strategy\cite{yamauchi1997frontier} to identify areas in the partially-explored map to reduce the occupancy map entropy, i.e., converting unknown cells into free or obstacle. A frontier is a contiguous set of unknown cells that are all directly adjacent to free cells. Using a wavefront search\cite{keidar2012robot} over the occupancy map emanating from the robot pose and exploiting the contiguous property of frontier cells, we can find all frontiers in the partially-explored map. We produce a viewpoint for each frontier by computing a pose that obeys a number of constraints, e.g., away from obstacles but within sensor range. Since there are normally several frontiers, we assign a cost to each frontier viewpoint using Equation~\ref{eq:frontier}.

\begin{align}\label{eq:frontier}
    \mathcal{C}(V; R, G) = L(R,V) + \beta A(R, V) + \zeta\mathbb{I}(V \neq G)
\end{align}
where $V$ represents the viewpoint-in-question; $R$ is the current pose of the robot; and $G$ is the previous viewpoint. $L(R,V)$ measures the path distance from the robot pose to the viewpoint $V$ using the A${}^\star$ global planner. For non-holonomic robots, we also have an angular penalty $A(R,V)$, along with a weight $\beta$, that discourages the robot from selecting viewpoints that require turning around since those are expensive operations for non-holonomic robots. The final term $\mathbb{I}(V \neq G)$, along with its penalty $\zeta$, is an indicator function that returns $1$ if the viewpoint-in-question $V$ and previous viewpoint $G$ are not the same. This prevents the robot from oscillating between two similarly-priced viewpoints by penalizing changing the viewpoint.

We take the greedy approach and travel to the frontier with the lowest cost. The nature of exploration constantly causes the frontier to be pushed back and deformed; furthermore, localization drift and loop closures can shift the map causing frontiers to change as well. For this reason, we re-evaluate all frontiers at a fixed rate to ensure we have the latest frontiers and viewpoints.

\subsection{Lighthouse Definition and Creation}
As the robot explores an environment, SLAM provides it with pose estimates that drift along the length of the trajectory. To correct this drift, we use purely vision-based loop closures. Since the creation of a loop closure constraint requires comparing the visual features of the current keyframe with a previous one, more correspondences help in both creation and disambiguation of loop closure constraints.

\begin{figure}[h]
\centering
\includegraphics[width=\columnwidth]{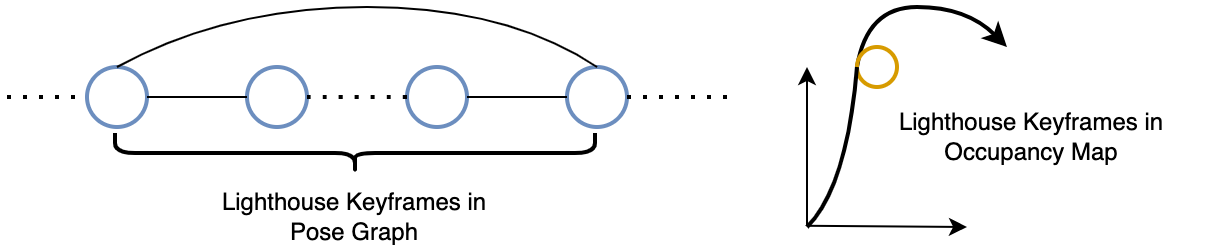}
\caption{\small{\textbf{Lighthouse.} In the pose graph, a lighthouse looks like a normal set of keyframes (blue circles) encompassed in a loop closure, but, in the occupancy map, that set of keyframes is generated by an in-place rotation which creates a little circle of spatially-clustered keyframes in the trajectory (orange circle).}}
\label{fig:lighthouse}
\end{figure}

For a narrow-FoV robot, we can \textit{emulate} a panoramic view by performing an in-place rotation. This is equivalent to having a ${360}^\circ$ view at the point where the robot stopped; intuitively, this is like ``looking back'' along the trajectory. But for narrow-FoV robots, this ``looking back'' requires extra motion and time so it would not be efficient to do this all across the map. Instead, while performing FE, we opportunistically and proactively identify a few key places to perform this in-place rotation. We monitor the incoming keyframes and, for each, we compute a \textbf{view score} that is the number of detected features in the keyframe. In addition to the panoramic view, the number of features also correlates to a higher likelihood of loop closure. If the view score is above a particular threshold, then we know that, if we were to perform an in-place rotation, then we have at least one view that is high in view score to encourage a likely loop closure. In the pose graph, this creates a special structure of keyframes that are both spatially and topologically clustered together, with a loop closure across all constituent keyframes. We call this cluster a \textbf{lighthouse} (Figure \ref{fig:lighthouse}). If, while performing FE, we detect an incoming keyframe that satisfies the view score threshold, we pause FE and create a lighthouse which will be used by the LH-ALC planner for loop closure.

\subsection{Lighthouse-based Active Loop Closure (LH-ALC)}

The purpose of ALC is to bound the pose uncertainty during FE by creating loop closure constraints in the pose graph. For keyframe-based vSLAM, this means traveling back to reacquire a view from some previous keyframe to trigger loop closure creation. However, for even smaller environments, vSLAM can produce hundreds, if not thousands, of keyframes that could be potential loop closure candidates. Furthermore, attempting to loop close at any one particular keyframe is very challenging since noise or drift may shift that keyframe making it difficult to reacquire. 

To remedy these problems, instead of traveling back to a particular keyframe, we travel to a \textit{lighthouse} and perform an in-place rotation. This greatly increases our likelihood of loop closure since the panoramic view detects many features that have a wider dispersion over the local environment than the features of a single keyframe. We call this flavor of ALC \textbf{Lighthouse-based Active Loop Closure (LH-ALC)}.

There are several mechanisms to trigger LH-ALC such as (i) a periodic timer and (ii) computing the relative uncertainty of latest keyframe from the nearest lighthouse. For the former, we trigger LH-ALC first after some time $T_0$ and keep scaling the timer by a factor $\gamma$ for every time we trigger LH-ALC: $T_0, \gamma T_0, \gamma^2 T_0,\cdots$. This ensures we aren't spatially bounded to some radius from any particular lighthouse. While this approach requires only constant-time evaluation, it can fail if the pose uncertainty has increased drastically between LH-ALC timers. The latter approach is more reactive to the pose uncertainty in that we only trigger LH-ALC when the relative uncertainty is high; however, computing the pose uncertainty of the latest keyframe is an expensive operation, requiring covariance estimation.

We unconditionally create the very first lighthouse, i.e., the \textbf{home lighthouse}, around the starting point of exploration. The home lighthouse is reliable because it is created near the origin so we collect additional panoramic views around the home lighthouse to ensure we can easily obtain a loop closure here. Note that since the home lighthouse is the starting point of exploration, a loop closure at the home lighthouse encompasses almost the entirety of the SLAM pose graph and is considered a global loop closure.

An additional graph strengthening measure we use is to connect each lighthouse to a nearby one. When we opportunistically create a lighthouse, instead of continuing FE immediately, we find the nearest lighthouse and travel to it to perform an in-place rotation. In the pose graph, this creates a loop that encompasses \textit{both} lighthouses, thereby connecting them to each other and lowering both of their relative uncertainties. Using this strategy, we effectively connect each lighthouse to the home lighthouse transitively. By performing this sequence of connections, we create even more loop closure constraints in the pose graph that encompass other loops; this creates a very stable mesh-like structure that helps bound the SLAM uncertainty. After connecting each new lighthouse to a previously-created one, we continue FE.

\subsection{Global Graph Stabilization (GGS) Planning}

The loop closures constraints created by the LH-ALC planner ensure local pose graph stability since we travel back to lighthouses or connect adjacent ones together. \textbf{Global Graph Stability (GGS)} is stability over the entire pose graph rather than just the subgraphs that encompass the adjacent lighthouse connections. The GGS planner creates an even more stable pose graph by analyzing regions of the map that are view-deficient and extending the trajectory along those regions \textit{in both directions} and connecting them to the home lighthouse. Our lighthouse creation and connection strategy of SAE, paired with the greedy approach of FE, effectively creates a spoke-and-hub pattern in the keyframes across the occupancy map where the ``spokes'' aren't necessarily connected. Furthermore, partial observability of the map during FE often interferes with the ability to connect the ``spokes''.

\begin{figure}
    \centering
    \begin{subfigure}[t]{0.45\columnwidth}
        \centering
        \includegraphics[width=\columnwidth]{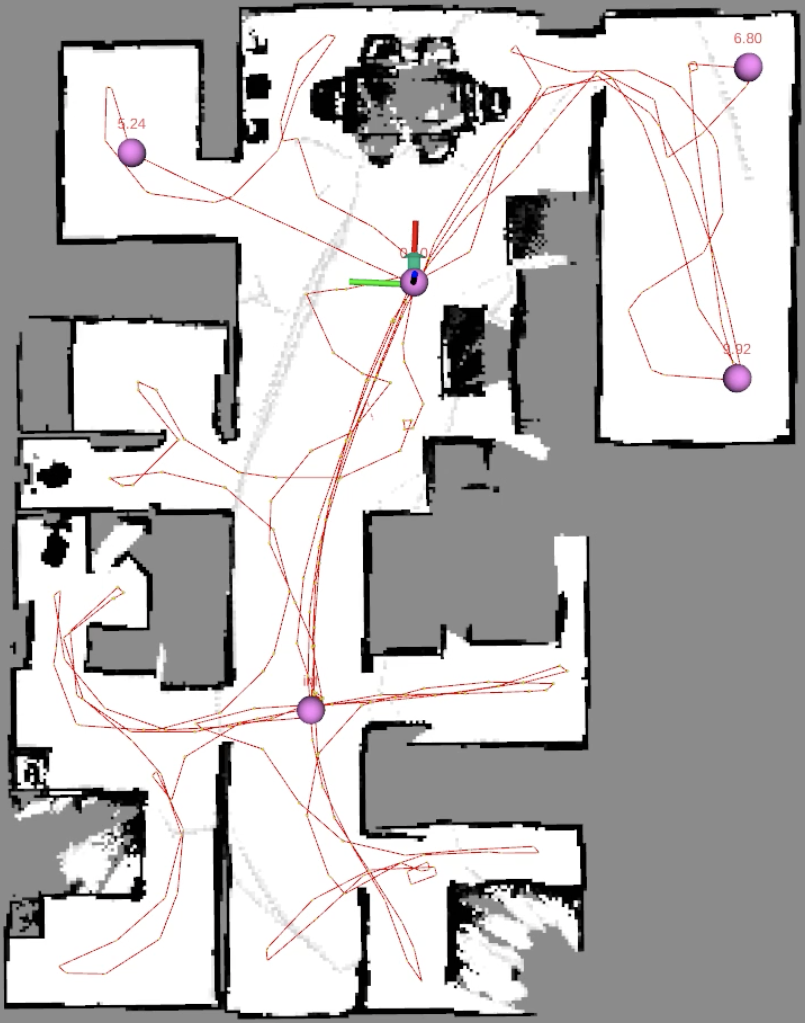}
        \caption{\small{\textbf{Pose Graph Before GGS.}}}
    \end{subfigure}
    \hfill
    \begin{subfigure}[t]{0.45\columnwidth}
        \centering
        \includegraphics[width=\columnwidth]{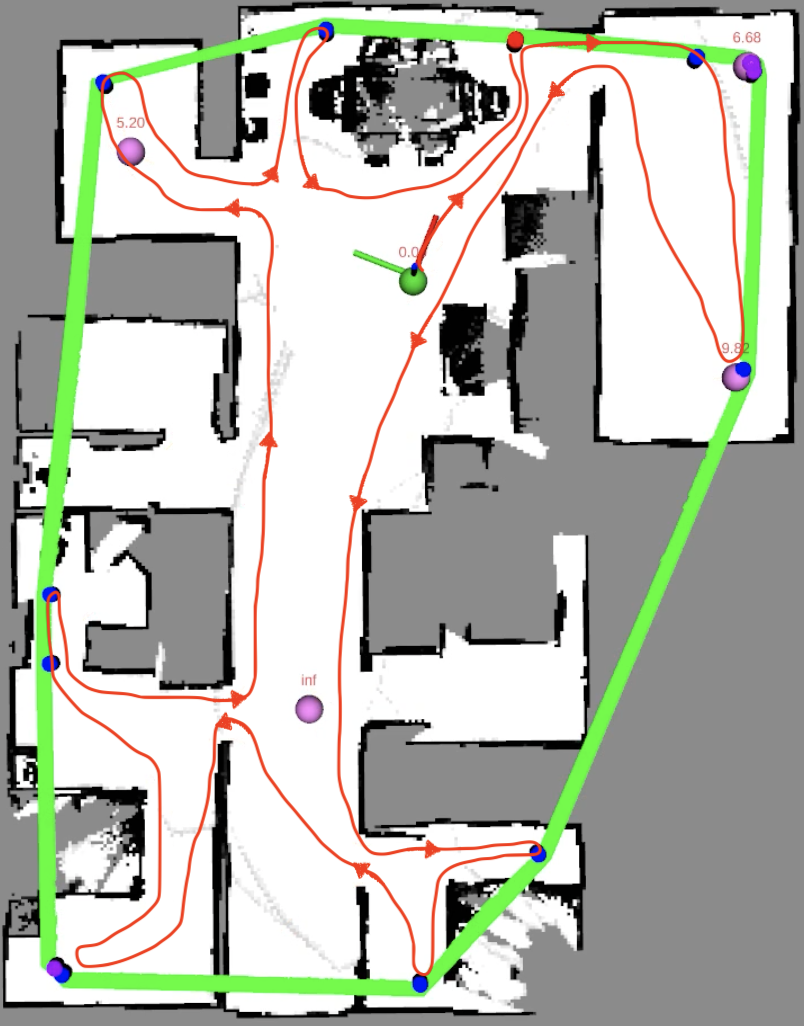}
        \caption{\small{\textbf{GGS Convex Hull.}}}
    \end{subfigure}
    \caption{\small{\textbf{Global graph stabilization (GGS).} The nature of our exploration strategy creates a spoke-and-hub pattern in the pose graph emanating from the lighthouses. This means the spokes are not necessarily connected at their farthest extents; they are only connected at the hub, i.e., lighthouse. The GGS planner consumes the keyframes or lighthouses, computes a convex hull (green lines), and travels to the vertices (blue circle; red is active) both clockwise and anti-clockwise to collect bi-directional views. For optimization, we skip vertices that are already close to others (purple circle). The red trajectory (with arrows denoting one directionality) is an example trajectory the robot would take and will be traversed both forwards and backwards.}}
    \label{fig:ggs}
\end{figure}

We wait until FE is completed and we have a completed occupancy map and locally-stable pose graph around the lighthouses. Performing global uncertainty calculations on the pose graph itself is often very expensive, but we can exploit this ``spoke-and-hub'' pattern in a much cheaper way. By taking all of the keyframes in the pose graph and computing the convex hull via the QuickHull algorithm\cite{barber1996quickhull}, the vertices become the endpoints of the ``spokes''. The GGS planner will travel to each of these vertices both clockwise and anticlockwise along the hull to connect the spokes and collect both forward and backward views to connect to the home lighthouse. Figure \ref{fig:ggs} shows the convex hull and path the device would take to each vertex. GGS planning connects even more views of the environment to the home lighthouse to ensure global stability of the pose graph. This stability creates more loop closure constraints when performing subsequent navigation on the explored map.

\subsection{System Design}

\begin{figure}[h]
\centering
\includegraphics[width=\columnwidth]{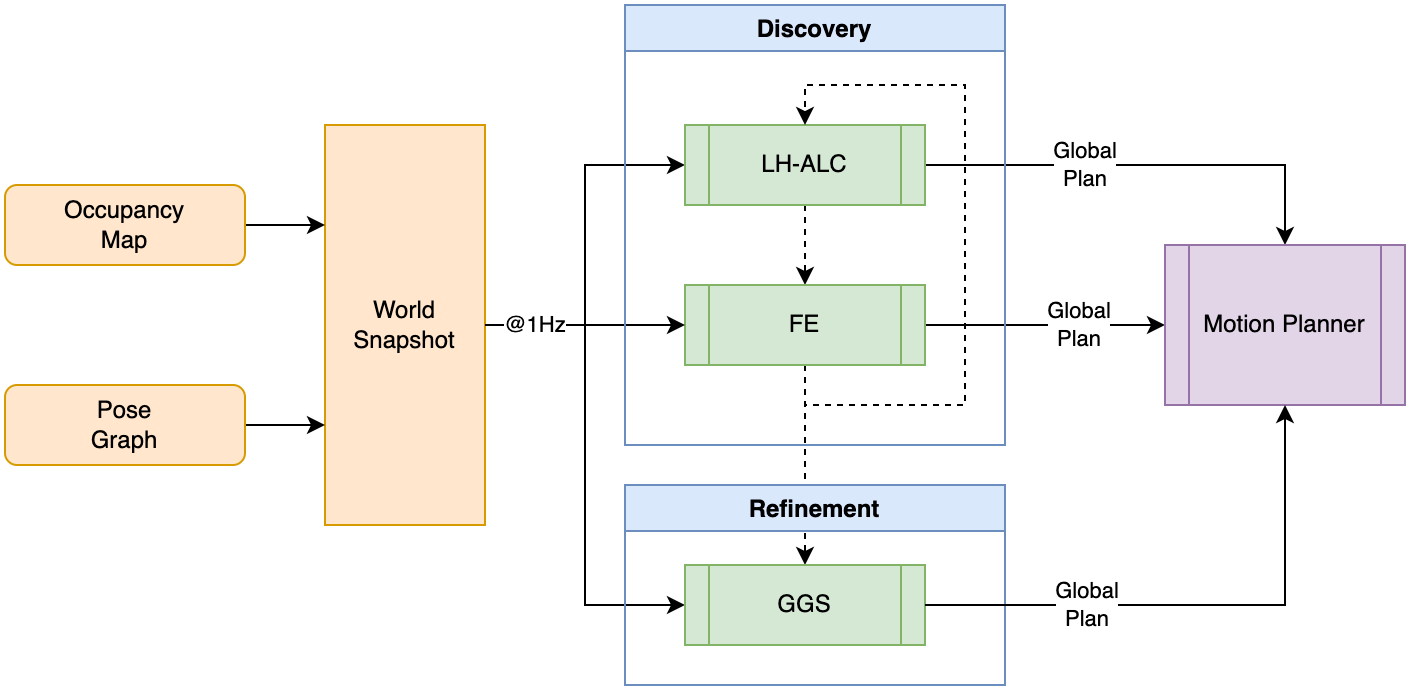}
\caption{\small{\textbf{System Design.} At a fixed rate, we collect perception data such as the occupancy map and pose graph and bundle it into the world snapshot that is accessible to all planners. We first start with the discovery phase and alternate between LH-ALC and FE. After those are finished, we move on to the refinement phase where we run the GGS planner. All planners issue plans to the motion planner.}}
\label{fig:sys-design}
\end{figure}

SAE has multiple planners that interact with each other. Figure \ref{fig:sys-design} illustrates our system design. Certain planners may preempt others or depend on them to finish operations. To ensure no contention between planners, we propose a system design based around planners that consume data from perception at a fixed rate, called the \textbf{world snapshot}, and produce plans to our motion planner. Furthermore, to enable GGS after FE, we define two phases of SAE: (i) discovery and (ii) refinement. For the former phase, we run two planners in order: LH-ALC and FE. LH-ALC starts by creating the home lighthouse. At the start of exploration, the SLAM uncertainty is near zero since the device starts at the origin so LH-ALC yields to FE immediately. During FE, if we notice a good view score, LH-ALC preempts FE to create the lighthouse and connect it to a previous lighthouse. If we detect SLAM uncertainty crosses a threshold or the timer triggers, we also preempt FE and travel to the closest lighthouse to create a loop closure constraint. The first phase finishes when FE finishes, and we travel back to the home lighthouse.

After the first phase, we have a completed occupancy map and the robot is at the home lighthouse. For the next refinement phase, we focus on the GGS planner and creating global stability for subsequent navigation. We exclusively run GGS planning during the second phase to ensure a globally stable pose graph for subsequent navigation. After GGS planning, we navigate the robot back to the home lighthouse for a final global correction.

\section{Experiments}
We evaluated our SAE approach on a nonholonomic mobile ground robot with a low compute budget and narrow-FoV active depth sensor and stereo camera in various indoor home environments. We use a keyframe-based pose graph vSLAM\cite{dellaert2017factor}. For our baseline exploration algorithm, we used a frontier-based exploration planner\cite{yamauchi1997frontier} with the wavefront frontier detector\cite{keidar2012robot}. Our mobile ground robot used two Qualcomm QCS605 octocore ARM processors, each with 2GB RAM and with two cores running at 2.5GHz and six cores running at 1.7GHz. The indoor home environments consist of a single floor and vary from about 1000 ft${}^2$ to 3000 ft${}^2$.

\subsection{Exploration Success}
A successful exploration is one where we can map the entire unknown environment without incurring so much localization drift that we cannot travel to unexplored regions or the home lighthouse. Since the quality of the map is dependent on the quality of localization\cite{liu2018ice} and ALC creates loop closures to stabilize the map, we performed an ablation study to measure the effect of the LH-ALC planner on overall exploration success. We ran exploration in 6 indoor home environments; in each environment, we ran ten exploration trials (half FE and half SAE) for a total of 60 samples. The robot started and ended in the same location in every environment.

\begin{figure}[h]
\centering
\includegraphics[width=\columnwidth]{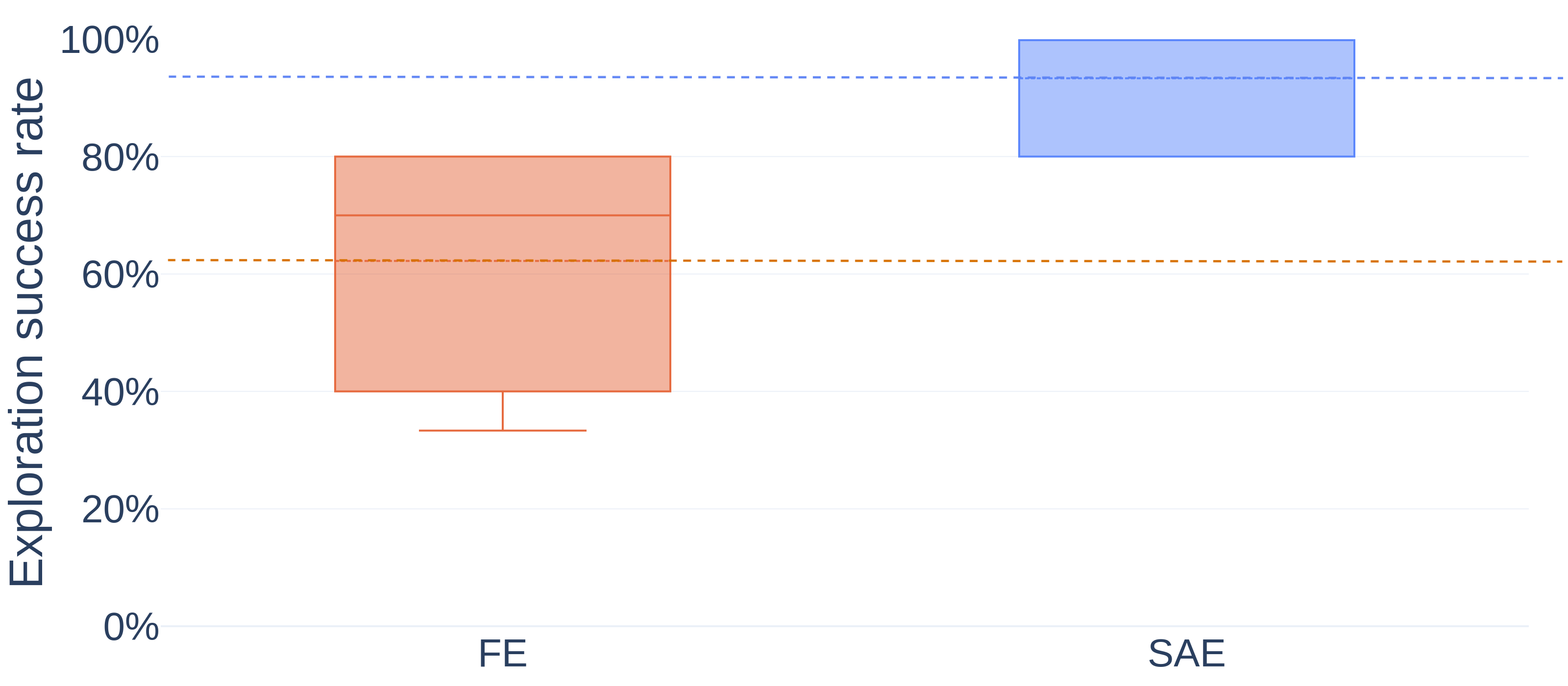}
\caption{\small{Exploration success rate for both FE and SAE in six indoor environments with five trials each. The mean success rate, represented by the dashed line, is 93\% for SAE compared to 62\% for FE. There's also a higher variance for FE indicating the success rate is left to chance of creating loop closures.}}
\label{fig:exp-success}
\end{figure}

%\begin{figure}[h]
%\centering
%\includegraphics[width=\columnwidth]{img/exploration_runtimes.png}
%\caption{\small{Exploration runtime increase distribution for SAE over FE. On average, SAE (without GGS) increases the runtime in an environment by a factor of 1.75 compared to FE.}}
%\label{fig:exp-runtimes}
%\end{figure}

\begin{figure}
\begin{tabular}{cccc}
\subfloat[FE home 1]{\includegraphics[width = 1.5in]{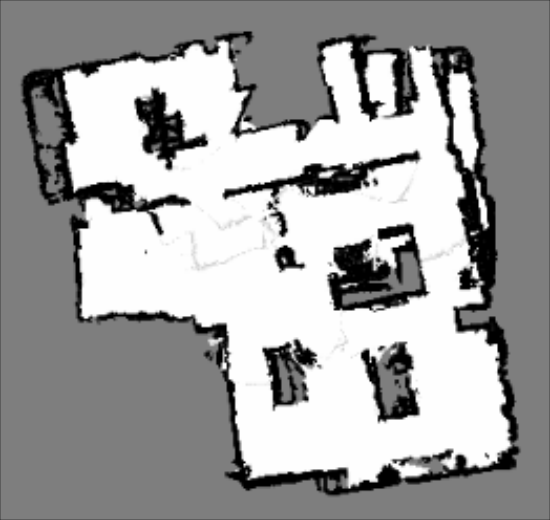}\label{subfig:fe-env-1}} &
\subfloat[SAE home 1]{\includegraphics[width = 1.5in]{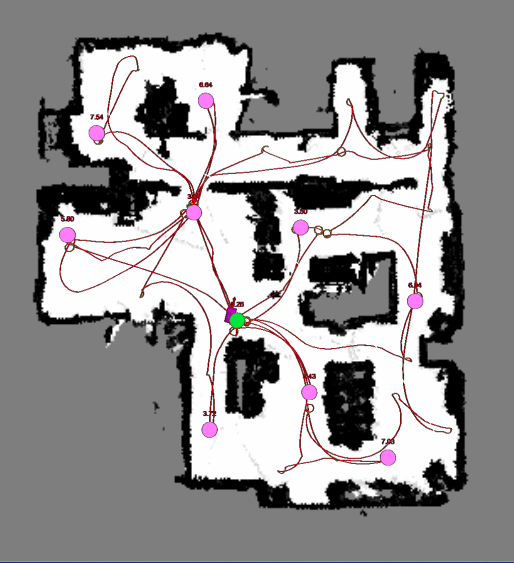}\label{subfig:alc-env-1}} \\
\subfloat[FE home 2]{\includegraphics[width = 1.5in]{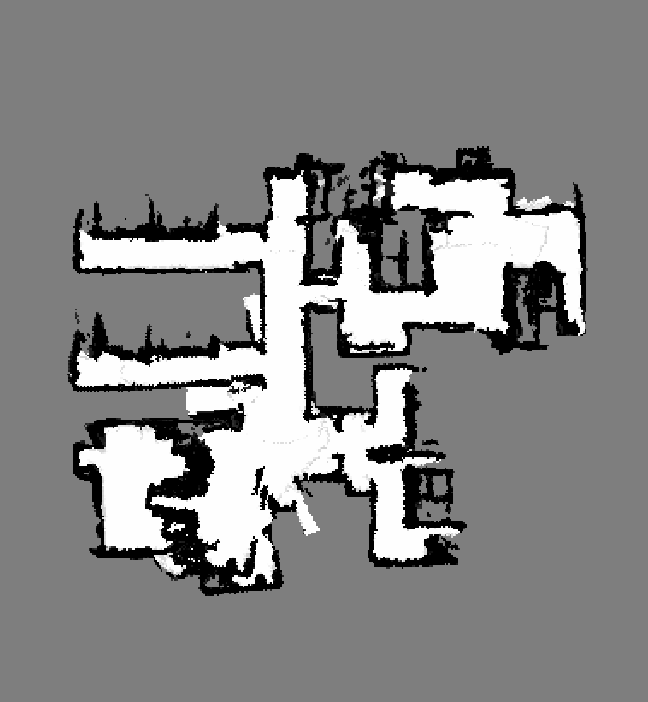}\label{subfig:fe-env-2}} &
\subfloat[SAE home 2]{\includegraphics[width = 1.5in]{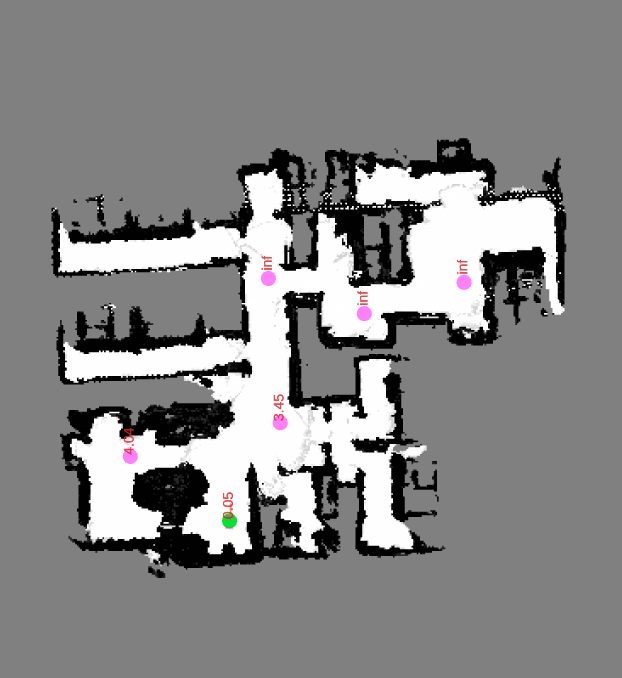}\label{subfig:alc-env-2}}
\end{tabular}
\caption{\small{Exploration maps in two homes comparing FE and SAE. Notice the map drift in the top right region of home 1 and bottom left region of home 2 with FE. These block off paths between the regions causing navigation failures. The maps on the right show the lighthouses and the resulting stable maps.}}
\label{fig:exp-maps}
\end{figure}

Figure \ref{fig:exp-success} shows the results averaged from several trials of each set. We see that SAE has a mean success rate of 93\% compared to FE with 62\%. On average, the SAE exploration time (without GGS) takes 1.75x longer than FE. The increase is due to the extra motion of actively creating more loop closures. Figure \ref{fig:exp-maps} shows example occupancy maps from two indoor home environments. We see map drifts with FE due to increasing pose uncertainty which causes paths to be blocked in the occupancy map. Exploration failures occur when the uncertainty of SLAM grows unbounded and distorts the occupancy map by placing obstacles in a way that blocks paths to prevent further exploration or returning to the home lighthouse. Such distortions can be found in the top right region in Figure \ref{subfig:fe-env-1} and bottom left region in Figure \ref{subfig:fe-env-2}. Since frontiers are created at the boundary of free and unknown space, FE has no incentive to travel to known spaces. However, SLAM benefits from traveling to previously observed areas since they are the only places to obtain loop closures. SAE keeps the pose uncertainty bounded during exploration which mitigates drift and allows exploration to complete successfully.

\subsection{Post-exploration Loop Closures}
Beyond exploration, subsequent navigation on the explored map should be successful as well and not produce drift. SAE has the GGS planner which globally stabilizes the pose graph to ensure the success of subsequent navigation. To understand the effectiveness of the GGS planner, we explore two real-world environments: a furnished and unfurnished home. We separately keep track of keyframes and loop closures during the discovery and refinement phases; since GGS is the only planner in the refinement stage, this becomes an ablation study assess how many additional keyframes and loop closures are created from the GGS planner to strengthen the pose graph.

\begin{table}
\centering
\resizebox{0.49\textwidth}{!}{
\begin{tabular}{ccc|cc}
& \multicolumn{2}{c}{Furnished Home} & \multicolumn{2}{c}{Unfurnished Home}\\
\hline
Total KFs & \multicolumn{2}{c}{5740} & \multicolumn{2}{c}{5086} \\
Total LCs & \multicolumn{2}{c}{161} & \multicolumn{2}{c}{121} \\
%Total explored area & \multicolumn{2}{c}{154$m^2$} & \multicolumn{2}{c}{172$m^2$} \\
%Total time & \multicolumn{2}{c}{1590s} & \multicolumn{2}{c}{1353s} \\
%Total KFs & \multicolumn{2}{c}{5740} & \multicolumn{2}{c}{5086} \\
\hline
& Discovery & Refinement & Discovery & Refinement\\
\hline
KFs created & 3814 & 1926 & 3688 & 1398\\
\% of total KFs & 66\% & 34\% & 73\% & 27\%\\
LCs created & 88 & 73 & 70 & 51\\
\% of total LCs & 55\% & 45\% & 58\% & 42\%\\
\hline
\end{tabular}}
\caption{\small{\textbf{Additional Keyframes (KFs) and Loop Closures (LCs) from GGS.} We separately track the number of KFs and LCs generated during each phase of exploration. We see that the GGS planner in the refinement stage creates significantly more KFs and LCs in addition to just the discovery phase.}}
\label{tab:kfs}
\end{table}

Table \ref{tab:kfs} shows the additional keyframes and loop closures generated by the GGS planner. We see that in both cases, the GGS planner contributes significantly to additional keyframes being added to the pose graph that provide even more views for subsequent navigation. The furnished home produces a slighly larger number of keyframes than the unfurnished one since the furniture has visual features on it and also creates occlusions. We also see that almost half of the total loop closures in the exploration run come from the GGS planner which globally stablizes the pose graph and reduces drift. 

We also compare the ability of the entire SAE against FE to improve subsequent navigation on the explored map. We explore the same environment twice, once with FE and another with SAE and save the maps. Then we travel through the same sequence of 50 fixed poses and measure the number of loop closures against the saved pose graph, which we call \textbf{relocalizations} instead of loop closures to distinguish the two.

\begin{figure}
\begin{tabular}{cccc}
\subfloat[FE]{\includegraphics[width = 1.6in]{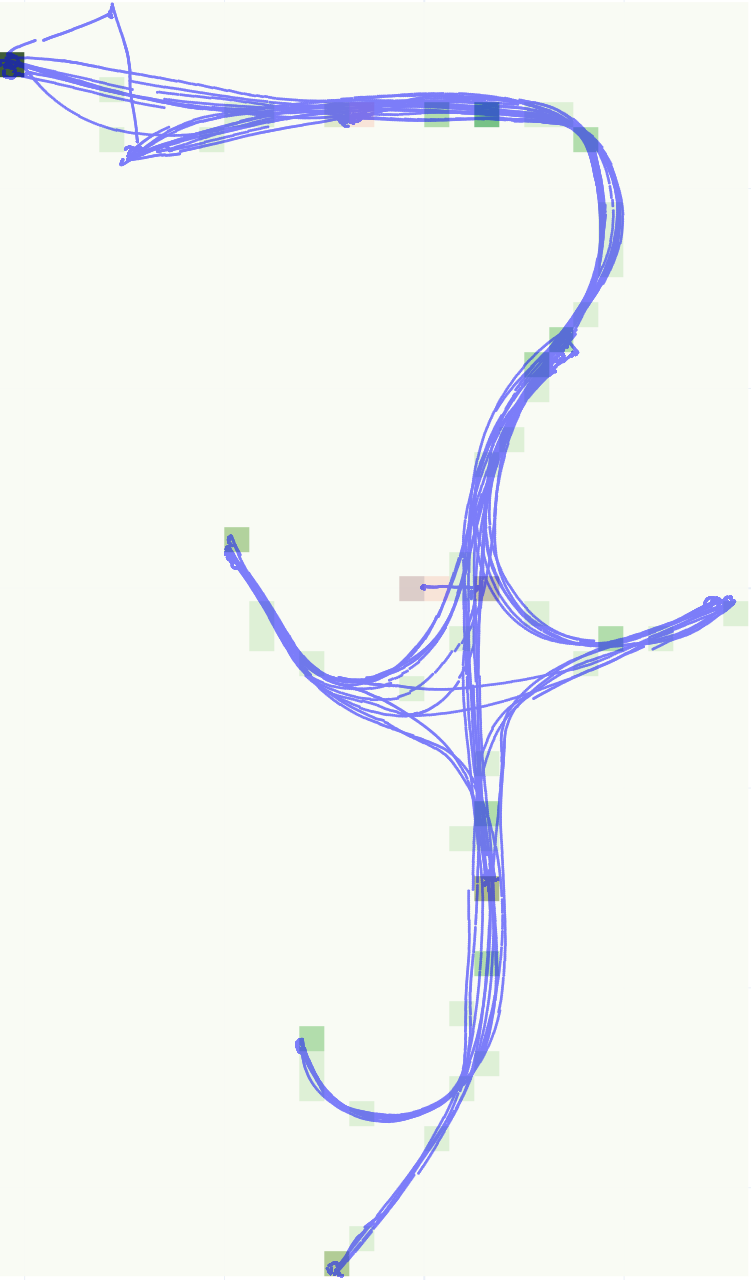}\label{subfig:reloc-fe}} &
\subfloat[SAE]{\includegraphics[width = 1.6in]{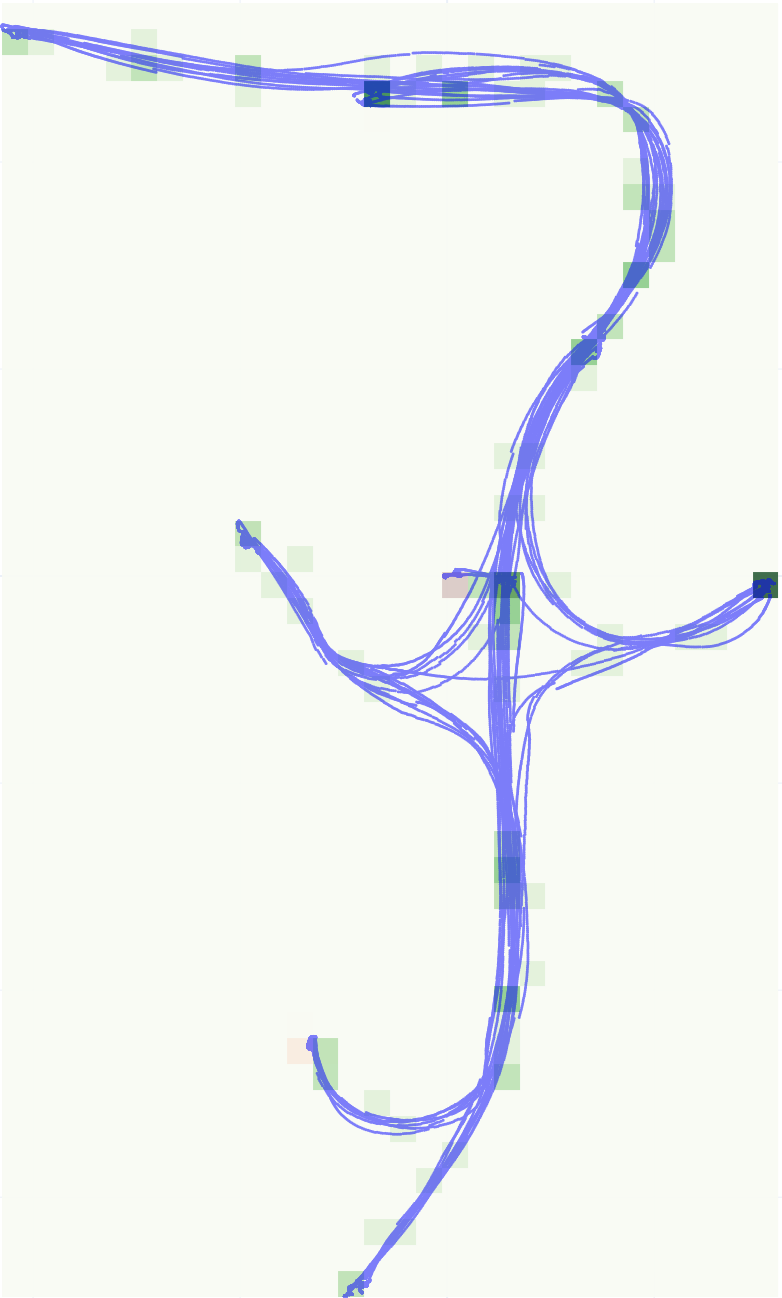}\label{subfig:reloc-lhalc}}
\end{tabular}
\caption{\small{\textbf{SAE's Effect on Post-exploration Navigation}. These relocalization heatmaps shows the cumulative trajectory of 50 navigation goals. Shaded green cells indicate the number of successful relocalization events in each of the discretized cells. Red colors indicate repeated relocalization events while the device was stationary to indicate any pauses between navigation goals. Figure \ref{subfig:reloc-lhalc} shows better trajectory estimation convergence, higher frequency of relocalizations, and a more diverse distribution of relocalization events.}}
\label{fig:reloc-heatmap}
\end{figure}

Figure \ref{fig:reloc-heatmap} shows the qualitative results comparing post-exploration navigation with FE and SAE. Quantitatively, the total number of SAE relocalizations was 315; FE achieved only 163 relocalization. There are almost twice the number of relocalizations for SAE. In general, observing more views of the environment increases the likelihood of relocalization which causes higher navigation success rates because the decreased pose uncertainty prevents occupancy map distortion. Views have directionality so observing views going in one direction of a path are different from views observed going in the opposite direction of the same path. The GGS planner directs the robot to observe views in both directions increase the overall likelihood of relocalization.

\section{Conclusion}
In this work, we presented a novel active SLAM exploration strategy that used lighthouses to ensure local and global SLAM pose graph stability. By creating, traveling back to, and connecting these panoramic views, we can ensure local graph stability through our LH-ALC planner. The GGS planner ensures global graph stability by connecting potentially nonadjacent views back to the home lighthouse.

In future work, we plan on investigating graph stability analysis techniques to better target which subgraphs and regions need the most focus during the refinement stage to reduce unnecessary motion to speed up exploration\cite{placed2022explorb}.


%%%%%%%%%%%%%%%%%%%%%%%%%%%%%%%%%%%%%%%%%%%%%%%%%%%%%%%%%%%%%%%%%%%%%%%%%%%%%%%%

%%%%%%%%%%%%%%%%%%%%%%%%%%%%%%%%%%%%%%%%%%%%%%%%%%%%%%%%%%%%%%%%%%%%%%%%%%%%%%%%

%%%%%%%%%%%%%%%%%%%%%%%%%%%%%%%%%%%%%%%%%%%%%%%%%%%%%%%%%%%%%%%%%%%%%%%%%%%%%%%%

\section*{ACKNOWLEDGMENT}
The authors would like to thank Chaitanya Desai, Rajasimman Madhivanan, Hesam Rabeti, Arnie Sen, Leena Vakil, and Roger Webster for their contributions to the ideas that went into this work.
%and for supporting the delivery of this work in the Consumer Robotics Department of Amazon Lab126.

%\clearpage
\bibliographystyle{IEEEtran}
\bibliography{refs}

\end{document}